\documentclass[letterpaper, 10 pt, conference]{ieeeconf}  

\IEEEoverridecommandlockouts                              

\usepackage{graphicx} 
\usepackage{amsmath} 
\usepackage{microtype}
\usepackage{todonotes}
\usepackage{units}
\usepackage{booktabs}
\usepackage{algorithm, algorithmic}
\usepackage{mathtools}
\usepackage[hidelinks]{hyperref}
\hypersetup{
    colorlinks=false,
    linkcolor=black,
    filecolor=black,      
    urlcolor=black
}
\usepackage{amssymb}

\usepackage[firstpage]{draftwatermark}
\SetWatermarkText{
\begin{minipage}{15cm}\centering
\vspace{-25cm}
\end{minipage}
}
\SetWatermarkScale{1}
\SetWatermarkLightness{0} 
\SetWatermarkFontSize{10pt}
\SetWatermarkAngle{0}

\usepackage{tikz}
\usepackage{pgfplots}
\usetikzlibrary{arrows,decorations,backgrounds,shapes,chains}
\pgfplotsset{compat=newest}
\pgfplotsset{plot coordinates/math parser=false}
\newlength\figureheight
\newlength\figurewidth

\usepackage{tabularx}

\makeatletter
\newcommand\footnoteref[1]{\protected@xdef\@thefnmark{\ref{#1}}\@footnotemark}
\makeatother

\newcommand{\vep}{\mbox{\boldmath $\epsilon$}}

\newcommand{\vta}{\mbox{\boldmath $\tau$}}

\newcommand{\vPh}{\mathbf \Phi}

\newcommand{\vd}{\mathbf d}
\newcommand{\ve}{\mathbf e}

\newcommand{\vg}{\mathbf g}
\newcommand{\vh}{\mathbf h}

\newcommand{\vl}{\mathbf l}

\newcommand{\vq}{\mathbf q}
\newcommand{\vr}{\mathbf r}
\newcommand{\vs}{\mathbf s}

\newcommand{\vu}{\mathbf u}

\newcommand{\vx}{\mathbf x}

\newcommand{\vdx}{\delta \mathbf x}
\newcommand{\vdu}{\delta \mathbf u}
\newcommand{\tdu}{\delta \bar {\mathbf u}}

\newcommand{\vA}{\mathbf A}
\newcommand{\vB}{\mathbf B}
\newcommand{\vC}{\mathbf C}
\newcommand{\vD}{\mathbf D}
\newcommand{\vE}{\mathbf E}

\newcommand{\vG}{\mathbf G}
\newcommand{\vH}{\mathbf H}

\newcommand{\vL}{\mathbf L}
\newcommand{\vM}{\mathbf M}
\newcommand{\vN}{\mathbf N}

\newcommand{\vP}{\mathbf P}
\newcommand{\vQ}{\mathbf Q}
\newcommand{\vR}{\mathbf R}
\newcommand{\vS}{\mathbf S}

\newcommand{\vU}{\mathbf U}

\newcommand{\tA}{\mathbf {\widetilde A}}
\newcommand{\tB}{\mathbf {\widetilde B}}

\newcommand{\tP}{\mathbf {\widetilde P}}
\newcommand{\tQ}{\mathbf {\widetilde Q}}
\newcommand{\tR}{\mathbf {\widetilde R}}

\newcommand{\tvg}{\mathbf {\widetilde g}}
\newcommand{\tq}{\widetilde q}
\newcommand{\tvq}{\mathbf {\widetilde q}}
\newcommand{\tvr}{\mathbf {\widetilde r}}

\newcommand{\bu}{\mathbf{\bar{u}}}

\newcommand{\hu}{\mathbf{\hat{u}}}
\newcommand{\hx}{\mathbf{\hat{x}}}

\newcommand{\cProj}{\mathcal P^n_{\mathcal N(\cdot)}}
\newcommand{\cProjfull}{\mathcal P^n_{
\mathcal N \left( \left[ \substack{\vE_n\\ \vM_n} \right] \right) 
}}

\title{\LARGE \bf
A Projection Approach to Equality Constrained Iterative Linear Quadratic Optimal Control$^\ast$
}

\author{Markus Giftthaler and Jonas Buchli$^\dagger$
\thanks{$^\dagger$Agile \& Dexterous Robotics Lab, Institute of Robotics and Intelligent Systems, ETH Z\"urich, Switzerland. {\small \{mgiftthaler, buchlij\}@ethz.ch}
$^\ast$Corrected version of the paper on \href{https://ieeexplore.ieee.org/document/8239538/}{\emph{IEEEXplore}}. Fixes a typo in Eq. (11)-(12).
}
}

\begin{document}
\maketitle
\thispagestyle{empty}
\pagestyle{empty}

\begin{abstract}
This paper presents a state and state-input constrained variant of the discrete-time iterative Linear Quadratic Regulator (iLQR) algorithm, with linear time-complexity in the number of time steps. The approach is based on a projection of the control input onto the nullspace of the linearized constraints. We derive a fully constraint-compliant feedforward-feedback control update rule, for which we can solve efficiently with Riccati-style difference equations. We assume that the relative degree of all constraints in the discrete-time system model is equal to one, which often holds for robotics problems employing rigid-body dynamic models. Simulation examples, including a 6 DoF robotic arm, are given to validate and illustrate the performance of the method.
\end{abstract}
%

\section{Introduction}
\label{sec:Introduction}
\subsection{Motivation}
In numerical optimal control for robotics, often also referred to as trajectory optimization, Differential Dynamic Programming (DDP)~\cite{mayne1966ddp} and its variants have recently received increased attention. In contrast to classical trajectory optimization approaches, such as Direct Collocation or Direct Multiple Shooting~\cite{diehl2006fast}, where open-loop trajectories are optimized, DDP-style methods are `optimal feedback planners', which design a stabilizing feedback and a feedforward controller at the same time. This family of algorithms solves optimal control problems including a nonlinear cost function and nonlinear system dynamics in an iterative way. Given an initial stabilizing control policy, a local linear approximation of the system dynamics and a linear-quadratic approximation of the cost function is computed. Then, an incremental improvement to the control law is designed and the process repeated until convergence is reached at a local minimum of the cost function. A major reason for the popularity of DDP-style methods is that their computational complexity is often linear in the number of time steps, $O(N)$. 

An important DDP variant is the iterative Linear Quadratic Regulator (iLQR)~\cite{todorov2005ilqg}, which is also known as
Sequential Linear Quadratic Optimal Control (SLQ)~\cite{slq:2005}.
While DDP is an exact-Hessian method requiring second-order derivatives of the system dynamics (note the exception in~\cite{unscented_ddp}), iLQR requires only first-order derivatives of the dynamics thanks to a Gauss-Newton Hessian approximation. This renders it an attractive choice for a large range of practical control applications, for example nonlinear model predictive control on robotic systems~\cite{neunert16hexrotor, giftthaler2017efficient, farshidian2017realtime}. iLQR has been applied for trajectory optimization of complex underactuated robots~\cite{neunert:2017:ral}, and has been combined with Machine Learning~\cite{mitrovic2010adaptive}.

To date, one of the main shortcomings of the algorithm concerns the treatment of constraints. So far, iLQR can handle bounds on the control input, see~\cite{tassa2014control}, with slight modifications. In~\cite{sideris2011riccati}, it is shown for the discrete-time case that state-control equality constraints can be incorporated elegantly through elimination of corresponding Lagrange Multipliers from the first-order necessary conditions of optimality. This approach maintains linear time complexity and results in a feedback-law respecting the state-input constraints. In the same work, an approach to treat pure state-constraints is presented, which however requires the solution of a large-scale linear system of equations, rendering the overall algorithm $O(N^3)$ and resulting in a feedback law which is not state-constraint compliant. The continuous-time counterpart for state-input equality constrained iLQR is given in~\cite{farshidian16efficient}. Additionally, pure-state equality and inequality constraints are incorporated as soft constraints through a penalty method. While the penalty method adds low computational overhead in each iLQR iteration, an increased number of overall iterations may be required due to the gradual `tightening' of the penalty coefficient. Furthermore, the resulting feedback gains are not compliant with the state constraints. 
Prior work in constrained DDP~\cite{Lin1991, xie2017differential} includes arbitrary nonlinear equality and inequality constraints, but requires solving Quadratic Programs for every time-step and every iteration in order to ensure strict feasibility during the forward integration.

In previously presented iLQR variants, the treatment of pure state constraints as `hard constraints' remained a bottleneck, both because of computational complexity and feedback. This paper contributes an additional perspective on equality-constrained iLQR and introduces a new approach in which the constraints are incorporated using a projection.

\subsection{Contribution}
\label{sec:contribution}
This paper presents a state and state-input hard-constrained iLQR variant. The algorithm is based on a projection of the control input onto the nullspace of the linearized constraints. While this ensures the system dynamics to evolve on a subspace tangent to the manifold of nonlinear constraints, the projection leads to a singular control weighting matrix in the linear-quadratic optimal control problem. By applying an existing result from the theory of singular optimal control, we derive a fully constraint-compliant feedforward-feedback control update rule, for which we can solve efficiently with Riccati-style difference equations with linear time complexity. Two simulation examples are given to validate and illustrate the performance of the approach.

\subsection{Outline}
This paper is structured as follows. In Section~\ref{sec:problem_statement}, the optimal control problem and its linear-quadratic approximation are introduced. In Section~\ref{sec:proj_slq}, we establish a way of addressing state and state-input constraints by means of a projection and derive the associated projected optimal control problem. We show a solution approach, derive the corresponding Riccati-like equations and present connections to existing work. Section~\ref{sec:constrained_sim} showcases two equality-constrained simulation examples from a robotics context. 
A discussion and outlook conclude the paper in Section~\ref{sec:Discussion}.

\section{Problem Statement}
\label{sec:problem_statement}
Consider the following discrete-time, finite-horizon, constrained nonlinear optimal control problem
\begin{align}
&\min_{\vu_n} \left \{\Phi(\mathbf x_{N})+\sum_{n=0}^{N-1} L_n (\mathbf x_n,\mathbf u_n, n) \right \}
\label{eq:nonlinear_cost}\\
&\textrm{subject to}  \notag \\
& \mathbf x_{n+1}=\mathbf f_n(\mathbf x_n,\mathbf u_n, n),\hspace{2mm} \mathbf x(0)=\mathbf x_0 
\label{eq:nonlinear_dynamics}\\
& \mathbf g_1(\vx_n, \vu_n, n)= 0 
\label{eq:nonlinear_state_input_constraint}\\
& \mathbf g_2(\vx_n, n) = 0 \hspace{12mm} \forall n\in \{0,\ldots N-1\} \label{eq:nonlinear_state_constraint}
\\
& \mathbf g_3(\vx_N, N) = 0 \label{eq:nonlinear_terminal_state_constraint}
\end{align}
with state-vector $\mathbf{x}_n \in \mathbb R^m$ and control input vector $\mathbf{u}_n \in \mathbb R^p$. $L_n$ denotes the intermediate, non-negative cost rate at time-step $n$ and $\Phi(\mathbf{x}_N)$ the terminal, non-negative cost at time-step~$N$. $\vg_1(\cdot)$ and $\vg_2(\cdot)$ are the nonlinear state-input and pure state intermediate equality constraints, $\vg_3(\cdot)$ is the terminal pure state constraint, respectively.

The locally-optimal control law is constructed in an iterative fashion. In each
iteration, starting at a constraint-satisfactory initial condition \mbox{$\vx(0)=\mathbf{x}_0$}, we obtain nominal state and control input trajectories $\{\hx_n\}$ and $\{\hu_n\}$ through a forward integration of the system using the current policy. Note that, in the first iteration, a stable initial control policy $\vu_{init}(\vx_n, n)$ is required.
Next, we linearize the system dynamics and constraints and quadratically approximate the cost around the nominal trajectories. Denoting state and control deviations as $\vdx_n = \vx_n-\hx_n$ and $\vdu_n = \vu_n - \hu_n$ gives rise to the following constrained, linear-quadratic optimal control problem
\begin{align}
\min_{\delta \vu_n} \bigg\{&q_N +\vdx_N^\top\vq_N+\tfrac{1}{2}\vdx_N^\top\vQ_N\vdx_N  \notag \\ 
	&+\sum_{n=0}^{N-1} q_n+ \vdx_n^\top\vq_n + \vdu_n^\top \vr_n+ \tfrac{1}{2}\vdx_n^\top\vQ_n\vdx_n \notag  \\
	& \quad +\tfrac{1}{2}\vdu_n^\top\vR_n\vdu_n + \vdu_n^\top\vP_n\vdx_n \bigg\} 
	\label{eq:linearized_cost}\\
&\textrm{subject to} \notag \\ 
&\delta \vx_{n+1} = \vA_n \delta\vx_n+ \vB_n \delta\vu_n \label{eq:linear_dynamics} \\
&\vD_n \delta \vx_n + \vE_n \delta\vu_n = \ve_n  \label{eq:linear_state_input_constraint}\\
&\vC_n \delta \vx_n = \vd_n \hspace{12mm} \forall n\in \{0,\ldots N-1\} \label{eq:linear_state_constraint}
\\
&\vC_N \delta \vx_N = \vd_N \label{eq:linear_terminal_state_constraint}
\end{align} 
where $q$, $\vq$, $\vQ$, $\vr$, $\vR$ and $\vP$ are the coefficients of the Taylor expansion of the cost function~\eqref{eq:nonlinear_cost} around the nominal trajectories, Equation~\eqref{eq:linear_dynamics} is the linearized system dynamics and Equations~\eqref{eq:linear_state_input_constraint} to~\eqref{eq:linear_terminal_state_constraint} are the linearized constraints. 
We assume $\vR_n > 0$, and $\vQ_n\textrm{,} \  \vQ_N \geq 0$.
The goal is to solve this constrained linear-quadratic subproblem for a local control policy update $\vdu_n (\vx_n, n)$ and to update the overall control policy accordingly, \mbox{$\vu_n(\vx_n,n) \leftarrow \vu_n(\vx_n,n) + \vdu_n(\vx_n, n)$}.

\section{Projected Iterative Linear Quadratic Optimal Control}
\label{sec:proj_slq}
\subsection{Projecting Constraints}
Besides assuming that the problem is not over-constrained, we require Equations~\eqref{eq:linear_state_input_constraint} to~\eqref{eq:linear_terminal_state_constraint} to be well-defined in the following sense. First we assume, without loss of generality, that the state-input constraint matrix~$\vE_n$ has full row rank. If this condition is violated for some $n \in \{0,\ldots N-1\}$, the linearized state-input constraint needs to be factorized into a pure state constraint which can be appended to Equation~\eqref{eq:linear_state_constraint} and a remaining state-input constraint with full rank. Second, we assume that all $\vC_n$ and $\vC_N$ are full row rank (otherwise, remove linearly dependent rows).
Third, we assume that the relative degree $r_n$ of the constrained system is well-defined and $r_n = 1$ $\forall n \in \{0, 1, \ldots, N-1 \}$. Consider the matrices $\vM_n$, $\vN_n$ defined as follows
\begin{align}
\vM_n &= \vC_{n+1} \cdot \vB_n \ \textrm{,} \label{eq:M_r} \\
\vN_n &= \vC_{n+1} \cdot \vA_n \label{eq:N_r}
\end{align}
which quantify the propagation of a state-input pair $\vdu_n$,~$\vdx_n$ through the linear system dynamics, into the linear state constraint equation at the next time step. The constrained system given by Equations~\eqref{eq:linear_dynamics} to~\eqref{eq:linear_terminal_state_constraint} is said to have relative degree one at time index $n$, if the matrices $\vM_n$ and $\vC_{n+1}$ are both full row rank. 

In this work, we account for the linearized constraints in the optimal control problem~\eqref{eq:linearized_cost}-\eqref{eq:linear_terminal_state_constraint} as follows. By means of a projection, we limit the set of admissible control inputs~$\vdu_n$ at time index $n$ to such controls which
\begin{itemize}
    \item lead to a satisfaction of the linearized state-input constraints immediately and
    \item lead to satisfaction of the linearized pure state constraints at the next upcoming timestep.
\end{itemize}
Hence, we rewrite the pure state constraint in terms of a `previewed' state-input constraint
\begin{equation}
\left[ \substack{\vD_n\\ \vN_n} \right] \vdx_n 
+ \left[ \substack{\vE_n\\ \vM_n} \right] \vdu_n 
= \left[ \substack{\ve_n\\ \vd_{n+1}} \right]
\label{eq:total_constraint}
\end{equation}
and at time index $n$, the control input is to be selected such that it complies with Equation~\eqref{eq:total_constraint}, while also minimizing the cost function~\eqref{eq:linearized_cost}.
To achieve both, we propose the following structure for the control update $\vdu_n$
\begin{equation}
    \vdu_n = \vdu_n^0 + \cProjfull \cdot \bu_n
    \label{eq:control_ansatz}
\end{equation}
with $\cProj \in \mathbb R^{p\times p}$ being a projection matrix onto the nullspace of $\left[\vE_n^\top \ \vM_n^\top \right]^\top$ (for readability, we use the short notation $\cProj$ in the following).
While $\bu_n$ can be freely selected in order to control the system in the nullspace of the linearized constraints, we dedicate $\vdu_n^0$ to controlling the system \emph{onto} this nullspace in the first place. By inserting Equation~\eqref{eq:control_ansatz} into~\eqref{eq:total_constraint} and forming an unweighted Moore-Penrose pseudo-inverse, we derive the $L^2$-optimal choice for~$\vdu_n^0$
\begin{equation}
\vdu_n^0 =
\underbrace{\left[ \substack{\vE_n\\ \vM_n} \right]^\dagger \left[ \substack{\ve_n\\ \vd_{n+1}} \right] }_{=: \ \vep_n}
\underbrace{-\left[ \substack{\vE_n\\ \vM_n} \right]^\dagger  \left[ \substack{\vD_n\\ \vN_n} \right] }_{=: \ \vU_n}
\vdx_n \ 
\label{eq:choice_for_u0}
\end{equation}
Note that it decomposes into a feedforward term $\vep_n$, which can be interpreted as the minimum corrective action to eliminate constraint violations $\ve_n$ and $\vd_{n+1}$, and a feedback matrix $\vU_n$ which governs a possibly deviating state back onto the constraint nullspace.

In the following, we write $\vdu_n^0$ generally as \mbox{$\vdu_n^0 = \vep_n + \vU_n \vdx_n$} and the complete control increment correspondingly as 
\begin{equation}
\vdu_n = \vep_n + \vU_n \vdx_n + \cProj \bu_n
\label{eq:affine_input}
\end{equation}

In order to compute the pseudo-inverse within $\vep_n$ and $\vU_n$ we perform a matrix factorization of $\left[\vE_n^\top \ \vM_n^\top \right]^\top$ in terms of an SVD or a QR decomposition at every time step $n \in \{0, \ldots, N-1\}$. Note that as a side-product of such a decomposition, we can obtain a set of basis vectors for the nullspace of $\left[\vE_{n}^\top \ \vM_{n}^\top \right]^\top$, required to compute the projection matrix $\mathcal P^{n}_{\mathcal N(\cdot)}$.

\subsection{Reformulation as Singular Optimal Control Problem}
Rewriting the linearized system dynamics~\eqref{eq:linear_dynamics} using the restructured input~\eqref{eq:affine_input} leads to a new, affine system equation
\begin{align}
&\delta \vx_{n+1} = \tA_n \delta \vx_n+ \tB_n \delta \bu_n + \tvg_n \hspace{4mm} \textrm{with} \notag \\
&\tA_n = \vA_n+\vB_n \vU_n \hspace{3mm} \tB_n = \vB_n\cProj  \hspace{3mm}  \tvg_n = \vB_n \vep_n \label{eq:projected_system_matrices}
\end{align}

Similarly, it allows us to reformulate the original linear-quadratic optimal control problem~\eqref{eq:linearized_cost}-\eqref{eq:linear_terminal_state_constraint} into a projected optimal control problem with respect to $\delta \bu_n$
\begin{align}
\min_{\delta \bu_n} \bigg\{&q_N +\vdx_N^\top\vq_N+\tfrac{1}{2}\vdx_N^\top\vQ_N\vdx_N  \notag \\ 
	&+\sum_{n=0}^{N-1} \tq_n+ \vdx_n^\top\tvq_n + \tdu_n^\top \tvr_n+ \tfrac{1}{2}\vdx_n^\top\tQ_n\vdx_n \notag  \\
	& \quad +\tfrac{1}{2}\tdu_n^\top\tR_n\tdu_n + \tdu_n^\top\tP_n\vdx_n \bigg\} \label{eq:cost_function_tilde}
	\\
\textrm{s.t.} \hspace{2mm} 
&\delta \vx_{n+1} = \tA_n \delta \vx_n+ \tB_n \delta \bu_n + \tvg_n \label{eq:affine_system}
\end{align} 
where we define the projected weighting matrices for the intermediate cost as
\begin{align}
&\tq_n =  q_n + \vep_n^\top \vr_n + \tfrac{1}{2} \vep_n^\top \vR_n \vep_n \label{eq:qv_tilde}\\
&\tvq_n =  \vq_n + \vU_n^\top \vr_n + \vP_n^\top \vep_n + \vU_n^\top \vR_n \vep_n\\
&\tQ_n =  \vQ_n + \vU_n^\top \vR_n \vU_n + \vU_n^\top \vP_n + \vP_n^\top \vU_n \\
&\tvr_n =  \cProj (\vr_n + \vR_n \vep_n) \\
&\tR_n = \cProj \vR_n \cProj \\
&\tP_n = \cProj (\vP_n + \vR_n \vU_n)  \label{eq:P_tilde}
\end{align}
Note that the terminal cost remains unchanged.
The intermediate cost remains quadratic in~$\vdx_n$ and~$\tdu_n$. $\tQ_n$ remains positive semi-definite. However, the new input cost weighting matrix $\tR_n$ may become non-negative definite (singular) for any originally positive-definite matrix $\vR_n$ due to the multiplication with the constraint null-space projector~$\cProj$.
Hence, in order to compute the optimal control update $\tdu_n^*$ which minizes the cost-to-go for time~$n$, we need to solve a \emph{singular optimal control} problem.

\subsection{Computing the Optimal Control Increment}
Assume a quadratic value function of the form
\begin{equation}
\label{eq:cost_to_go}
V_{n}(\delta \mathbf x_{n}) = s_{n} + {\delta \mathbf x_{n}^\top \mathbf s_{n}} 
 + \tfrac{1}{2} \delta \mathbf x_{n}^\top \mathbf S_{n} \delta \mathbf x_{n}
\end{equation}
with weighting matrices $\vS_n \in \mathbb R^{m \times m}$, $\vs_n \in \mathbb R^{m \times 1}$ and $s_n \in \mathbb R$. From a value function perspective, the optimal control update can be derived by minimizing $V_n$ for a given $\vdx_n$. To the best of our knowledge, classical iLQR approaches have so far only treated the case of a strictly positive-definite Hessian approximation, equivalent to
\mbox{$\vB_n^\top \vS_{n+1} \vB_n + \vR_n > 0$},
which does not directly transfer to the optimal control problem~\eqref{eq:cost_function_tilde}-\eqref{eq:affine_system}.
In this work, however, we exploit the fact that for any discrete-time linear-quadratic optimal control problem, despite its possible singularity, the associated Riccati Difference Equation is well-defined, as long as a certain set of conditions is met~\cite{clements1978singular}.
For the projected optimal control problem~\eqref{eq:cost_function_tilde}-\eqref{eq:affine_system}, these conditions can be interpreted as 
\begin{align}
&\tB_n^\top \vS_{n+1} \tB_n + \tR_n \geq 0 \quad \textrm{and} \label{eq:condition_singular_hessian_1} \\
&\mathcal N(\tB_n^\top \vS_{n+1} \tB_n + \tR_n) \subset \mathcal N(\tA_n \vS_{n+1} \tB_n + \tP_n^\top) \ \textrm{.} \label{eq:condition_singular_hessian_2}
\end{align}
For a detailed discussion and derivation of such conditions, the interested reader is referred to the original work by Clements and Anderson in~\cite{clements1978singular}.
Since we assume that $\vR_n > 0$, it is straight-forward to show that Equations~\eqref{eq:condition_singular_hessian_1} and~\eqref{eq:condition_singular_hessian_2} are satisfied for any symmetric, positive semi-definite weighting matrix $\vS_{n+1}$.
Hence, the singular optimal control problem~\eqref{eq:cost_function_tilde}-\eqref{eq:affine_system} can be addressed using the well-established quadratic value-function approach, where the Hessian's inverse can be replaced by its Moore-Penrose pseudo-inverse. 

It is simple to show that, if the value function~\eqref{eq:cost_to_go} is quadratic in $\vdx_{n+1}$ for time index $n+1$, it remains quadratic during back-propagation in time, given the affine system dynamics in Equation~\eqref{eq:affine_system} and the linear quadratic cost in Equation~\eqref{eq:cost_function_tilde}. According to Bellman's Principle of Optimality, the optimal control input $\tdu_n^*$ at the preceding time index $n$ arises from
\begin{align*}
&V^*_n(\vdx_n) = \min_{\tdu_n} \bigg[ \tq_n + \vdx^\top (\tvq_n + \tfrac{1}{2} \tQ_n \vdx_n) 
+ \tdu_n^\top \tP_n \vdx_n  \notag \\
&+ \tdu_n^\top(\tvr_n + \tfrac{1}{2} \tR_n \tdu_n)
+ V^*_{n+1}(\tA_n \vdx_n + \tB_n \tdu_n +\tvg_n) \bigg]
\end{align*}
Inserting the value function~\eqref{eq:cost_to_go} together with the affine system dynamics~\eqref{eq:affine_system} and minimizing the overall expression w.r.t. $\tdu_n$ leads to an optimal control update of the form
\begin{equation}
\label{eq:optimal_control_input}
\tdu_n^*= -\vH_n^\dagger \vh_n - \vH_n^\dagger \vG_n \vdx_n
\end{equation}
where we have defined the abbreviations
\begin{align*}
\vh_n & = \tvr_n + \tB_n^\top (\vs_{n+1} + \vS_{n+1} \tvg_n )\notag\\
\mathbf G_n & = \tP_n + \tB_n^\top \vS_{n+1} \tA_n \notag\\
\mathbf H_n & = \tR_n + \tB_n^\top \vS_{n+1} \tB_n \ \textrm{.}
\end{align*}
Note that $\vH_n^\dagger$ denotes the unweighted Moore-Penrose pseudo-inverse of the possibly singular Hessian approximation $\vH_n$. By plugging the optimal control update~\eqref{eq:optimal_control_input} into the above expression for $V^*_n(\vdx_n)$ we can confirm the quadratic nature of the optimal value function in $\vdx_n$. 

After equating coefficients with a quadratic value function Ansatz of form~\eqref{eq:cost_to_go} for $\vdx_n$, we define 
$\vl_n = -\vH_n^\dagger \vh_n$ 
and 
$\vL_n = - \vH_n^\dagger \vG_n$
and obtain the following, recursive Riccati-like difference equations for $\vS_n$, $\vs_n$ and $s_n$
\begin{align}
\vS_n &= \tQ_n + \tA_n^\top \vS_{n+1} \tA_n - \vL_n^\top \vH_n \vL_n \label{eq:riccati_vS}\\
\vs_n &= \tvq_n + \tA_n^\top (\vs_{n+1} + \vS_{n+1} \tvg_n) \notag \\& \qquad + \vG_n^\top \vl_n + \vL_n^\top (\vh_n + \vH_n\vl_n)\\
s_n &= \tq_n + s_{n+1} + \tvg_n^\top \vs_{n+1} + \tfrac{1}{2} \tvg_n^\top \vS_{n+1} \tvg_n \notag \\
& \qquad + \vl_n^\top (\vh_n + \tfrac{1}{2} \vH_n \vl_n) \label{eq:riccati_s}
\end{align}
which are valid for all $n \in {0,\ldots, N-1}$. For the final time-step~$N$ we obtain the terminal conditions \mbox{$\vS_N = \vQ_N$}, \mbox{$\vs_N = \vq_N$} and \mbox{$s_N = q_N$}.

Eventually, after combining Equations~\eqref{eq:affine_input} and~\eqref{eq:optimal_control_input}, the incremental controller update becomes
\begin{equation*}
\vdu_n = \vep_n + \vU_n \vdx + \cProj (\vl_n + \vL_n \vdx)
\end{equation*}
or after slight reformulation
\begin{equation}
\vdu_n (\vx_n) = \vep_n + \cProj \vl_n + (\vU_n + \cProj \vL_n )(\vx_n-\hx_n) \ \textrm{.}
\label{eq:new_control_update_law}
\end{equation}

\subsection{Interpretation and Connection to Existing Results}
Examining Equation~\eqref{eq:new_control_update_law}, it becomes obvious that the control update consists of a feedforward and a feedback-term, which are both composed of one part stemming from the computation of $\vdu_n^0$ and one part from the projected iLQR iteration performed in subspaces tangent to the constraint manifold. At this point, several connections to existing results in the field can be made:
\begin{itemize}
\item Considering the overall feedback matrix \mbox{$\vU_n + \cProj \vL_n$}, there is an obvious connection to the work by Hemami et al. in~\cite{hemami1979modeling}, in which it is shown that the set of all constraint compliant feedback matrices for a pair of linear system matrices $\vA, \vB$ is an affine subset of all admissible, linear state-variable feedback gains. In our case, this affine subset is offset from the origin by $\vU_n$ and spanned by the basis of the constraint nullspace, as indicated by $\cProj$.
\item Special case 1: when only pure state-constraints of form $\vC \vx_n=0$ are considered for a steady setpoint, our iLQR-iteration simplifies to the state-constrained LQR formulation and corresponding Riccati-like equations as presented by Ko et al. in~\cite{ko2005optimal}.
\item Special case 2: when only state-input constraints are considered, our iLQR-iteration simplifies to the result for state-input constraints given by Sideris et al. in~\cite{sideris2011riccati}.
\item The presented approach can be interpreted as a variant of hierarchical DDP~\cite{romano2015prioritized}, with a first task forming the constraints and a second task encoded in the cost.
\end{itemize}

\subsection{Main Iteration}
Algorithm~\ref{alg:projected_slq} summarizes the main iLQR iteration. The final control update after each Riccati backward pass is obtained through a line search where we select an $\alpha \in [0,1]$ such that the control policy
\begin{equation*}
\hu_n + \alpha \cdot (\vep_n + \cProj \vl_n) + (\vU_n + \cProj \vL_n )(\vx_n-\hx_n)
\end{equation*}
minimizes a merit function $\mathcal J$ defined as
\begin{equation}
\mathcal J = \Phi(\mathbf x_{N})+\sum_{n=0}^{N-1} \left( L_n (\mathbf x_n,\mathbf u_n, n) + \sigma \cdot  \lvert \left[ \substack{\ve_n\\ \vd_{n+1}} \right] \rvert \right) \ \textrm{.}
\label{eq:merit_function}
\end{equation}
The merit function consists of the non-projected cost function, the real positive tuning parameter $\sigma$ and the $L^1$-norm of the equality constraint violation evaluated for every $\alpha$.
We consider the algorithm converged when both the relative merit function change and the constraint's integrated square error (ISE) are below some user-defined threshold.
\begin{algorithm}[tpb] 
\caption{Equality Constrained iLQR Algorithm} 
\label{alg:projected_slq}
\begin{algorithmic} \scriptsize \STATE \textbf{Given} 
\STATE - Nonlinear dynamics, cost function and constraints as given in Equations~\eqref{eq:nonlinear_cost}-\eqref{eq:nonlinear_terminal_state_constraint}
\STATE - Initial stable control law, $\mathbf{\vu}_{init}(\vx,n)$ 
\STATE - merit function weighting parameter $\sigma$
\REPEAT 
\STATE \textbf{Forward Pass}
\STATE - forward simulate the system~\eqref{eq:nonlinear_dynamics}, obtain nominal trajectories
\STATE \hspace{1em}$\hx_0,\hu_0,\hx_1,\hu_1,\dots,\hx_{N-1},\hu_{N-1},\hx_N$
\STATE - Quadratize cost function along the trajectory to obtain LQ approximation~\eqref{eq:linearized_cost}
\STATE - Linearize the system dynamics and constraints along the trajectory to \STATE \hspace{1em}obtain the linear approximations in Equations~\eqref{eq:linear_dynamics},~\eqref{eq:linear_state_input_constraint},~\eqref{eq:linear_state_constraint} and~\eqref{eq:linear_terminal_state_constraint}
\STATE - perform SVD or QR decomposition of every $\left[\vE_n^\top \ \vM_n^\top \right]^\top$ and generate a 
\STATE \hspace{1em}set of basis vectors for its nullspace in order to compute $\vep_n$, $\vU_n$ as well 
\STATE \hspace{1em}as $\cProj , \ \forall n=0,1,\ldots,N-1$
\STATE \textbf{Backward Pass}
\STATE - compute the projected system- and weighting matrices in in~\eqref{eq:projected_system_matrices}, \eqref{eq:qv_tilde}-\eqref{eq:P_tilde}
\STATE - backwards solve the Riccati-like difference equations~\eqref{eq:riccati_vS}-\eqref{eq:riccati_s} with 
\STATE \hspace{1em}boundary conditions $\vS_N = \vQ_N$, $\vs_N = \vq_N$ and $s_N = q_N$
\STATE \textbf{Line search}
\STATE  search  over $\hu_n + \alpha \cdot (\vep_n + \cProj \vl_n) + (\vU_n + \cProj \vL_n )(\vx_n-\hx_n)$
\STATE - set $\alpha = 0$, integrate system, evaluate constraint errors and cost and 
\STATE \hspace{1em}compute reference merit function $\mathcal J_0$ according to Equation~\eqref{eq:merit_function}
\STATE - reset $\alpha = 1$
\REPEAT 
\STATE - Update the control,  forward simulate and compute new merit
\STATE - decrease $\alpha$ by a constant $\alpha_{d}$: $\alpha = \alpha / \alpha_{d}$
\UNTIL{found lower merit than $\mathcal J_0$ or number of max. line search steps reached}
\UNTIL{merit function converged \& $ISE < ISE_{max}$}
\end{algorithmic} 
\end{algorithm}

\section{Simulation Examples}
\label{sec:constrained_sim}
\subsection{Discretizing Continuous-time Rigid Body Dynamics}
\label{sec:discretizing_rbd}
While the presented constrained iLQR algorithm is formulated in discrete-time, most of the robotics modelling tools available today generate continuous-time ordinary differential equations. In this work, we auto-generate the dynamics and kinematics ODEs using the robotics code generator `RobCoGen'~\cite{frigerioCodeGen}, and obtain the continuous-time dynamics derivatives through algorithmic differentiation using our open-source framework~\cite{adrlCT,neunert:2016:derivatives}.
 
For time discretization, we assume that the control input is interpolated as zero-order hold.
Consider the example of a pure state constraint on position level where the continuous-time rigid body dynamics of a robot model equation reads as \mbox{$\vM(\vq) \ddot \vq + \vH(\vq, \dot \vq) + \vPh(\vq) = \vS \vta$}, with generalized coordinates $\vq$. 
In the theoretic case of using a simple explicit Euler integrator both for the forward rollout of the system dynamics and for the discretization of the system matrices $\vA_n$ and $\vB_n$, at least two time-steps would be required until the control input torque became `visible' to the position constraint, in terms of a change of $\vq$. Correspondingly, the system's relative degree would be at least two and the projected iLQR algorithm presented in this paper would not be admissible.

Therefore, for the following rigid-body system examples, the forward integration is performed using higher-order Runge-Kutta integrators. We obtain the system matrices $\vA_n$, $\vB_n$ through integrating a sensitivity ODE, which proved to be sufficient for rendering the systems considered in this paper relative degree 1 in various different simulation settings.
All simulations were done in C++.
\subsection{Multicopter with Body Position Constraints}
\begin{figure}
\centering
\includegraphics[width=0.99\columnwidth]{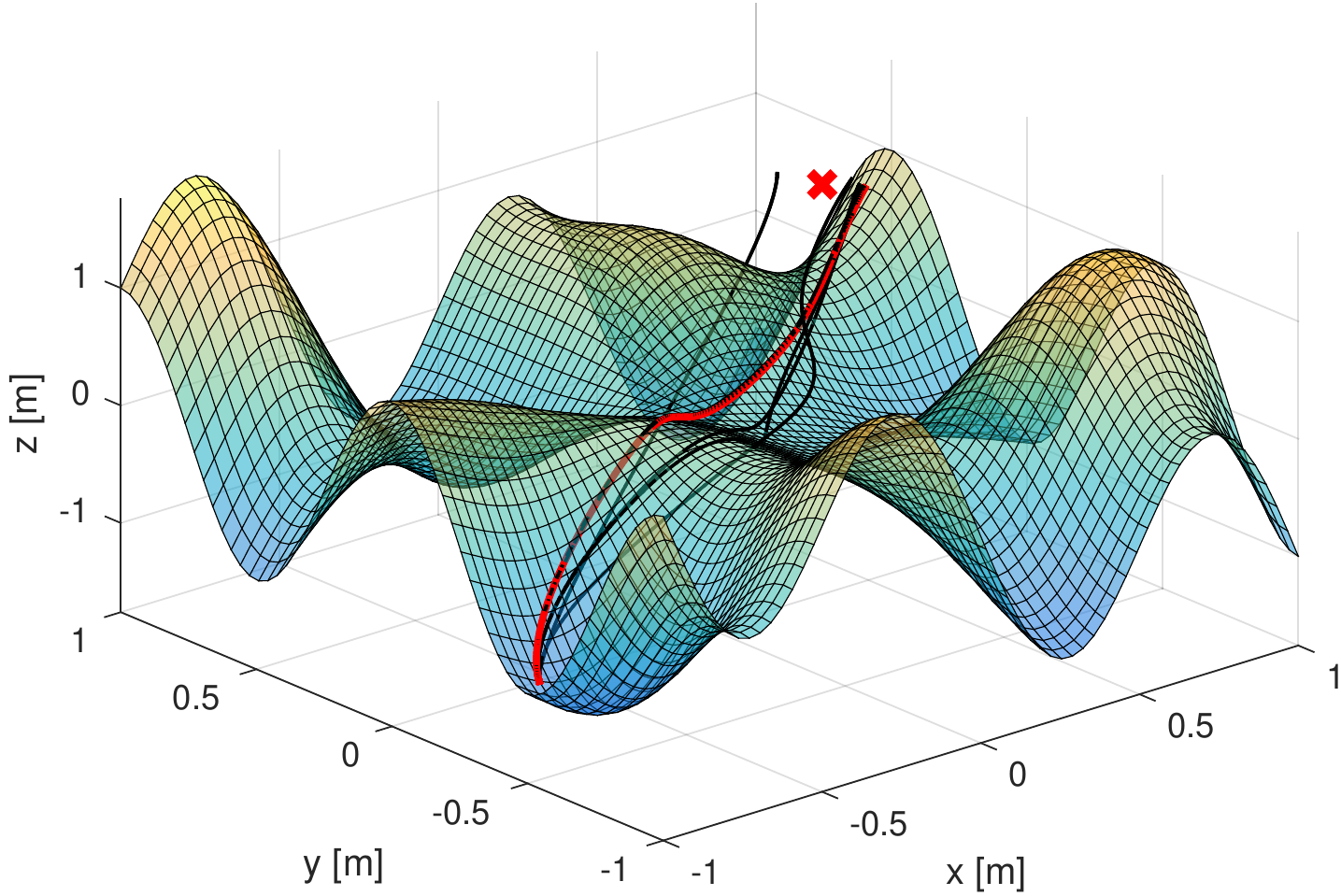}
\caption{This plot illustrates the nonlinear constraint landscape in the multicopter simulation example. The desired target position is marked with a cross and does not lie on the constraint surface. The optimal trajectory of the multicopter's CoM (red) ends at the point on the constraint surface closest to the target. The trajectories corresponding to intermediate iterates are shown in black.}
\label{fig:quadr_example}
\end{figure}

In a first example, we apply the algorithm to a multicopter trajectory optimization problem, where the system is modelled similar to~\cite{neunert16hexrotor}. In order to illustrate the algorithm's capability to deal with complex nonlinear equality constraints, we show a positioning task including position constraints on the multicopter Center of Mass (CoM). With the CoM position being expressed in $x,y,z$ coordinates, we consider the constraint \mbox{$y \sin(2\pi x) - x \cos(2 \pi y) - z = 0$}.
The multicopter needs to reach a desired position as closely as possible while respecting the constraint landscape. For a maneuver with time horizon $3$~sec, and time discretization $dt=10$~ms, the optimization converges in~7 iterations with a total constraint ISE $<10^{-3}$, and CPU time 0.14 sec (Intel Core i7-4850HQ CPU, 2.30~GHz). A plot of the trajectory is given in Fig.~\ref{fig:quadr_example}.

\subsection{6 DoF Robot Arm with End-Effector Position Constraint}
\begin{figure}
\centering
\includegraphics[width=0.99\columnwidth]{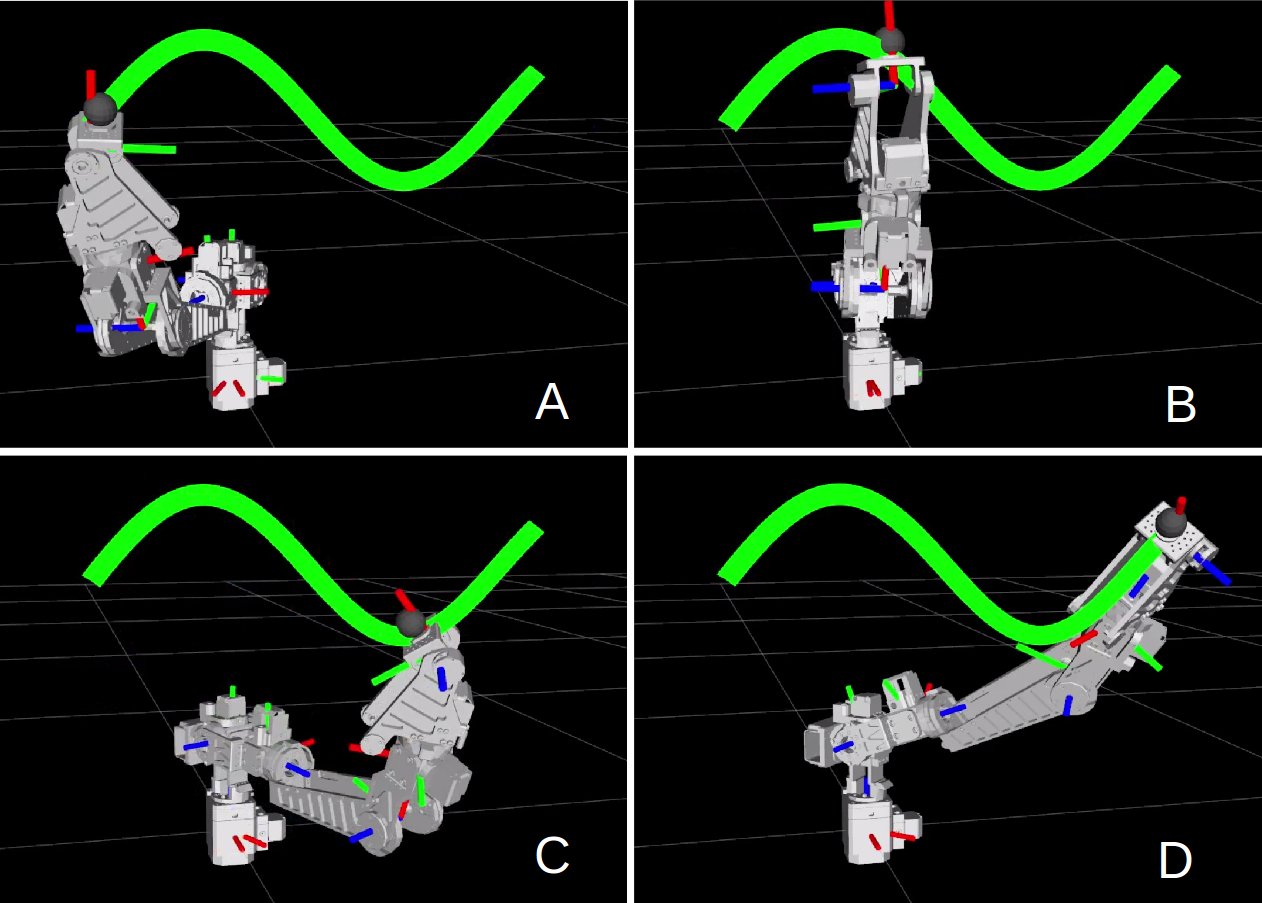}
\caption{Snapshots of an end-effector position constrained motion of a 6~DoF robot arm. `A' shows the initial pose, the sine-shaped end-effector position constraint is shown in green. Convergence was obtained after 10 iterations, with a total constraint ISE of $0.0013$, and a total CPU-time of 0.70 seconds. The time horizon was $3$~sec, the time discretization was chosen to be $5$~ms.}
\label{fig:arm_example}
\end{figure}

\begin{figure}[tbp]
    \centering
    \includegraphics[width=0.99\columnwidth]{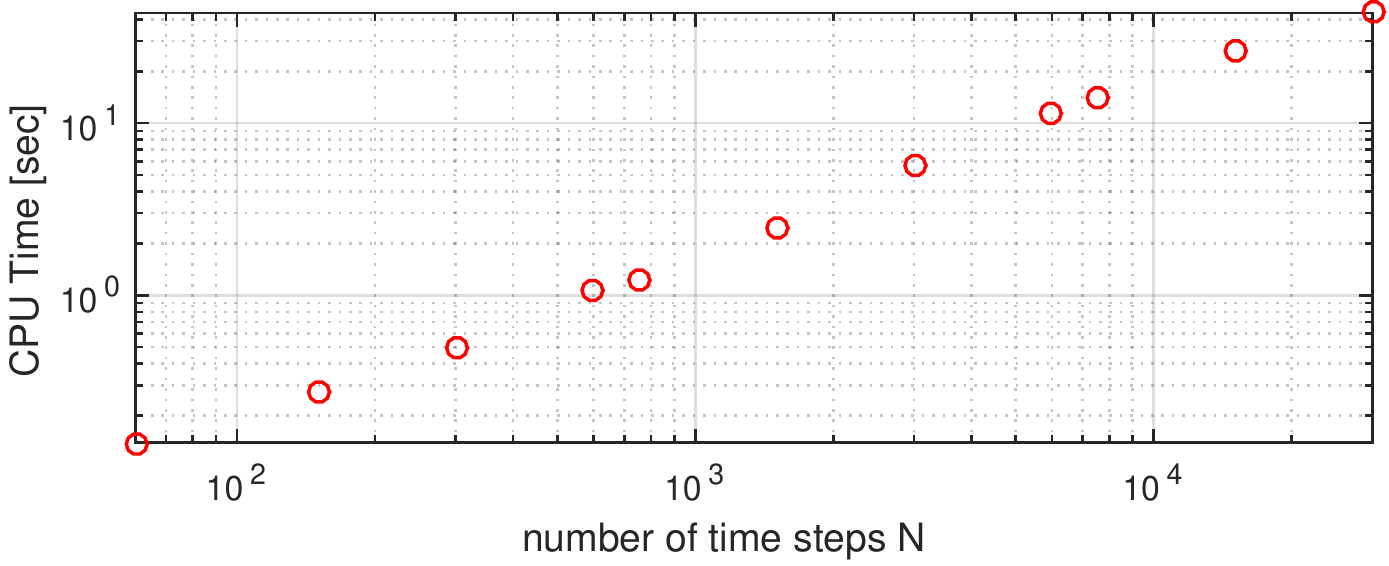}
    \caption{Average CPU times for the state-constrained robot arm task with different time discretizations. The time horizon was fixed at $3.0$~sec. Each data point represents the average of 100 runs with identical settings.}
    \label{fig:linear_time_comlexity}
\end{figure}

In this example, we showcase a trajectory optimization example for a fixed-base 6~DoF robot arm subject to an end-effector position constraint. The employed arm model is inspired by~\cite{phd16brehman}. Starting at an initial position `A', the task objective encoded in the cost function is to rotate the first joint about 90$^\circ$, while the end-effector path is constrained to lie on a sinusoid curve starting at the initial end-effector pose. The initial guess is a simple inverse-dynamics feedforward for the initial pose, combined with an infinite-horizon LQR feedback controller. Fig.~\ref{fig:arm_example} shows snapshots of the obtained trajectory. Fig.~\ref{fig:linear_time_comlexity} shows the average CPU-time for the same problem using different sampling times $dt$ and indicates linear time-complexity~$O(N)$.

\section{Discussion and Outlook}
\label{sec:Discussion}
This paper introduces a state and state-input constrained, projection-based iLQR variant, along with promising simulation results.
Essentially, the presented approach is a Single-Shooting Gauss-Newton type method. As such, it is crucial that a stable, initial control policy is provided. Note that this initial control policy is not required to lead to a constraint satisfactory trajectory (in fact, during the course of the simulation experiments, we observed that a wide range of different initial controllers and motions would lead to closely similar final solutions within a similar number of iterations). However, in order to achieve a sufficiently small overall constraint violation, the initial state $\vx_0$ is strictly required to be constraint compliant. Otherwise, it requires at least one time step with non-zero constraint violation to pass before the system can be steered onto the constraint manifold.

A unique feature of the presented algorithm is that it optimizes a feedforward trajectory, while also providing a control feedback that is guaranteed to be state-constraint compliant. While we have already used the state-input constrained feedback gains in hardware experiments in~\cite{giftthaler2017efficient} as a special case of this work, a verification of the purely-state constrained gains for practical robotics tasks is subject to ongoing work.

In terms of computational complexity, the algorithm is $O(N)$ in the number of time-steps $N$. In comparison to traditional unconstrained iLQR, the only computationally expensive operation added at each time step is the factorization of a stacked matrix reflecting state and state-input constraints.
It shall be noted that this process bears potential for computational parallelization, as the factorizations can be performed independently at every time step.

The method is free of tuning parameters, except for the cost function weights, which is the same for unconstrained iLQR, and for the merit function weight~$\sigma$.
In practice, the algorithm converges to the same solution for a broad range of choices for $\sigma$, in all shown simulation examples convergence was achieved with low tuning effort starting from rough guesses for the parameter.

A current limitation is certainly the restriction to systems with relative degree equal to 1. Essentially, the relative degree expresses after how many time steps of system evolution an initially state-input constraint compliant control at time index~$n$ becomes fully visible to the pure-state constraint. Note that an extension to relative degrees higher than one is possible, but does not easily generalize as it requires a case-specific examination of the linear independence constraint qualification of the constrained optimal control problem.
In certain cases, however, it may be possible that the constraints cannot be represented with relative degree one using the simple integration and discretization scheme from Section~\ref{sec:discretizing_rbd}, for example when complex actuator dynamics are included in the model. 
An extension of this work to cases with more complex constraints and dynamics is subject to future work. 
While initial simulation results look promising, another natural extension of this work is a rigorous comparison in terms of performance, convergence, robustness and computational complexity to reference implementations of established constrained optimization methods.

\section*{Acknowledgements}
\small
The authors would like to thank Moritz Diehl, Farbod Farshidian and Michael Neunert for fruitful discussions. This research was supported by the Swiss National Science Foundation through the NCCR Digital Fabrication and a Professorship Award to Jonas Buchli.

\bibliographystyle{ieeetr}
\bibliography{refs}

\begin{thebibliography}{10}

\bibitem{mayne1966ddp}
D.~Mayne, ``A second-order gradient method for determining optimal trajectories
  of non-linear discrete-time systems,'' {\em International Journal of
  Control}, vol.~3, no.~1, pp.~85--95, 1966.

\bibitem{diehl2006fast}
M.~Diehl, H.~G. Bock, H.~Diedam, and P.-B. Wieber, ``Fast direct multiple
  shooting algorithms for optimal robot control,'' in {\em Fast motions in
  biomechanics and robotics}, pp.~65--93, Springer, 2006.

\bibitem{todorov2005ilqg}
E.~Todorov and W.~Li, ``A generalized iterative {LQG} method for
  locally-optimal feedback control of constrained nonlinear stochastic
  systems,'' in {\em American Control Conference, 2005. Proceedings of the
  2005}, pp.~300--306, IEEE, 2005.

\bibitem{slq:2005}
A.~Sideris and J.~E. Bobrow, ``An efficient sequential linear quadratic
  algorithm for solving nonlinear optimal control problems,'' {\em Transactions
  on Automatic Control}, vol.~50, no.~12, pp.~2043--2047, 2005.

\bibitem{unscented_ddp}
Z.~Manchester and S.~Kuindersma, ``{Derivative-free trajectory optimization
  with unscented dynamic programming},'' in {\em IEEE 55th Conference on
  Decision and Control (CDC)}, pp.~3642--3647, dec 2016.

\bibitem{neunert16hexrotor}
M.~Neunert, C.~de~Crousaz, F.~Furrer, M.~Kamel, F.~Farshidian, R.~Siegwart, and
  J.~Buchli, ``Fast nonlinear model predictive control for unified trajectory
  optimization and tracking,'' in {\em IEEE International Conference on
  Robotics and Automation (ICRA)}, 2016.

\bibitem{giftthaler2017efficient}
M.~Giftthaler, F.~Farshidian, T.~Sandy, L.~Stadelmann, and J.~Buchli,
  ``{Efficient Kinematic Planning for Mobile Manipulators with Non-holonomic
  Constraints Using Optimal Control},'' in {\em IEEE International Conference
  on Robotics and Automation (ICRA)}, pp.~3411--3417, May 2017.

\bibitem{farshidian2017realtime}
F.~Farshidian, E.~Jelavi\'c, A.~Satapathy, M.~Giftthaler, and J.~Buchli,
  ``Real-time motion planning of legged robots: A model predictive control
  approach,'' in {\em 2017 IEEE-RAS 17th International Conference on Humanoid
  Robots (Humanoids)}, Nov 2017.

\bibitem{neunert:2017:ral}
M.~Neunert, F.~Farshidian, A.~W. Winkler, and J.~Buchli, ``Trajectory
  optimization through contacts and automatic gait discovery for quadrupeds,''
  {\em IEEE Robotics and Automation Letters (RA-L)}, 2017.

\bibitem{mitrovic2010adaptive}
D.~Mitrovic, S.~Klanke, and S.~Vijayakumar, ``Adaptive optimal feedback control
  with learned internal dynamics models,'' in {\em From Motor Learning to
  Interaction Learning in Robots}, pp.~65--84, Springer, 2010.

\bibitem{tassa2014control}
Y.~Tassa, N.~Mansard, and E.~Todorov, ``Control-limited {Differential Dynamic
  Programming},'' in {\em Robotics and Automation (ICRA), 2014 IEEE
  International Conference on}, pp.~1168--1175, IEEE, 2014.

\bibitem{sideris2011riccati}
A.~Sideris and L.~A. Rodriguez, ``A {Riccati} approach for constrained linear
  quadratic optimal control,'' {\em International Journal of Control}, vol.~84,
  no.~2, pp.~370--380, 2011.

\bibitem{farshidian16efficient}
F.~Farshidian, M.~Neunert, A.~W. Winkler, G.~Rey, and J.~Buchli, ``An efficient
  optimal planning and control framework for quadrupedal locomotion,'' in {\em
  IEEE International Conference on Robotics and Automation (ICRA)}, 2017.

\bibitem{Lin1991}
T.~C. Lin and J.~S. Arora, ``Differential dynamic programming technique for
  constrained optimal control,'' {\em Computational Mechanics}, vol.~9,
  pp.~27--40, Jan 1991.

\bibitem{xie2017differential}
Z.~Xie, C.~K. Liu, and K.~Hauser, ``{Differential Dynamic Programming with
  Nonlinear Constraints},'' in {\em IEEE International Conference on Robotics
  and Automation (ICRA)}, 2017.

\bibitem{clements1978singular}
D.~J. Clements and B.~D. Anderson, ``{Singular Optimal Control: The
  Linear-Quadratic Problem},'' in {\em Lecture Notes in Control and Information
  Science}, vol.~5, Springer-Verlag, Berlin, Germany, 1978.

\bibitem{hemami1979modeling}
H.~Hemami and B.~Wyman, ``Modeling and control of constrained dynamic systems
  with application to biped locomotion in the frontal plane,'' {\em IEEE
  Transactions on Automatic Control}, vol.~24, no.~4, pp.~526--535, 1979.

\bibitem{ko2005optimal}
S.~Ko and R.~R. Bitmead, ``Optimal control of linear systems with state
  equality constraints,'' {\em IFAC Proceedings Volumes}, vol.~38, no.~1,
  pp.~407--412, 2005.

\bibitem{romano2015prioritized}
F.~Romano, A.~Del~Prete, N.~Mansard, and F.~Nori, ``Prioritized optimal
  control: A hierarchical differential dynamic programming approach,'' in {\em
  2015 IEEE International Conference on Robotics and Automation (ICRA)},
  pp.~3590--3595, May 2015.

\bibitem{frigerioCodeGen}
M.~Frigerio, J.~Buchli, and D.~Caldwell, ``Code generation of algebraic
  quantities for robot controllers,'' in {\em Intelligent Robots and Systems
  (IROS), 2012 IEEE/RSJ International Conference on}, Oct 2012.

\bibitem{adrlCT}
M.~Giftthaler, M.~Neunert, M.~{St\"auble}, and J.~Buchli, ``{The {Control
  Toolbox} - An Open-Source {C++} Library for Robotics, Optimal and Model
  Predictive Control}.'' \url{https://adrlab.bitbucket.io/ct}, 2018.
\newblock arXiv:1801.04290 [cs.RO].

\bibitem{neunert:2016:derivatives}
M.~Neunert, M.~Giftthaler, M.~Frigerio, C.~Semini, and J.~Buchli, ``Fast
  derivatives of rigid body dynamics for control, optimization and
  estimation,'' in {\em 2016 IEEE International Conference on Simulation,
  Modeling, and Programming for Autonomous Robots (SIMPAR), San Francisco},
  pp.~91--97, Dec 2016.

\bibitem{phd16brehman}
B.~U. Rehman, {\em Design and Control of a Compact Hydraulic Manipulator for
  Quadruped Robots}.
\newblock PhD thesis, Istituto Italiano di Tecnologia (IIT) and University of
  Genova, 2016.

\end{thebibliography}

\end{document}